\documentclass[10pt, a4paper, copyright, logo, nonumbering]{instadeep}

\usepackage[utf8]{inputenc} % allow utf-8 input
\usepackage[T1]{fontenc}
\DeclareUnicodeCharacter{2032}{'}
\title{Multi-modal Transfer Learning \\ between Biological Foundation Models}

\correspondingauthor{jj.garauluis@instadeep.com ; g.richard@instadeep.com}

\author[*,1]{Juan Jose Garau-Luis}
\author[*,1]{Patrick Bordes}
\author[*,1]{Liam Gonzalez}

\author[1]{\\Masa Roller}
\author[1]{Bernardo P. de Almeida}
\author[2]{Lorenz Hexemer}
\author[2]{Christopher Blum}
\author[2]{Stefan Laurent}
\author[2]{Jan Grzegorzewski}
\author[2]{Maren Lang}

\author[**, 1]{\\Thomas Pierrot}
\author[**, 1]{Guillaume Richard}

\affil[*]{Equal contributions}
\affil[**]{Equal supervision}
\affil[1]{InstaDeep}
\affil[2]{BioNTech}

\usepackage{xspace}
\usepackage{amsmath}
\usepackage{amsfonts}       % blackboard math symbols

\newcommand{\xdna}{\mathbf{x}_{\text{dna}}}
\newcommand{\xrna}{\mathbf{x}_{\text{rna}}}
\newcommand{\xprot}{\mathbf{x}_{\text{prot}}}
\newcommand{\xrnai}{\mathbf{x}_{\text{rna}}^{(i)}}
\newcommand{\xproti}{\mathbf{x}_{\text{prot}}^{(i)}}
\newcommand{\expr}[1]{e\left(\ensuremath{#1}\right)}
\newcommand{\hdna}{\mathbf{h}_{\text{dna}}}
\newcommand{\hrna}{\mathbf{h}_{\text{rna}}}
\newcommand{\hprot}{\mathbf{h}_{\text{prot}}}

\newcommand{\hmm}{\mathbf{h_{\text{multi}}}}

\newcommand{\dnaalphabet}{\mathcal{A}_{\text{dna}}}
\newcommand{\rnaalphabet}{\mathcal{A}_{\text{rna}}}
\newcommand{\protalphabet}{\mathcal{A}_{\text{prot}}}
\newcommand{\thetadna}{\theta_{\text{dna}}}
\newcommand{\thetarna}{\theta_{\text{rna}}}
\newcommand{\thetaprot}{\theta_{\text{prot}}}

\definecolor{chineseblue}{rgb}{0.21,0.31,0.58}
\definecolor{myred}{rgb}{0.8,0,0}
\definecolor{mygreen}{rgb}{0,0.6,0}
\definecolor{myblue}{rgb}{0,0,0.7}
\definecolor{stanfordblue}{HTML}{006eb8}

\definecolor{myblue2}{rgb}{0,0,0.6}
\definecolor{mygreen2}{rgb}{0.2,0.6,0.2}
\definecolor{myred2}{rgb}{0.6,0,0}

\newcommand{\modelname}{IsoFormer }
\newcommand{\modelnamenospace}{IsoFormer}

\newcommand{\enformer}{Enformer\xspace}
\newcommand{\enformernospace}{Enformer}
\newcommand{\nt}{\textit{NT}\xspace}

\newcommand{\esm}{\textit{ESM}\xspace}

\newcommand{\dna}{\textit{DNA}\xspace}
\newcommand{\rna}{\textit{RNA}\xspace}
\newcommand{\dnanospace}{\textit{DNA}}
\newcommand{\rnanospace}{\textit{RNA}}

\newcommand{\mrna}{{\footnotesize m}\textit{RNA}\xspace}
\newcommand{\premrna}{{\footnotesize pre-m}\textit{RNA}\xspace}
\newcommand{\rnaseq}{\rnanospace{\footnotesize -seq}\xspace}

\usepackage{caption}

\begin{abstract}

Biological sequences encode fundamental instructions for the building blocks of life, in the form of \dnanospace, \rnanospace, and proteins. Modeling these sequences is key to understand disease mechanisms and is an active research area in computational biology. Recently, Large Language Models have shown great promise in solving certain biological tasks but current approaches are limited to a single sequence modality (\dnanospace, \rnanospace, or protein). Key problems in genomics intrinsically involve multiple modalities, but it remains unclear how to adapt general-purpose sequence models to those cases. In this work we propose a multi-modal model that connects \dnanospace, \rnanospace, and proteins by leveraging information from different pre-trained modality-specific encoders. We demonstrate its capabilities by applying it to the largely unsolved problem of predicting how multiple \rna transcript isoforms originate from the same gene (i.e. same \dna sequence) and map to different transcription expression levels across various human tissues. We show that our model, dubbed \modelnamenospace, is able to accurately predict differential transcript expression, outperforming existing methods and leveraging the use of multiple modalities. Our framework also achieves efficient transfer knowledge from the encoders pre-training as well as in between modalities. We open-source our model, paving the way for new multi-modal gene expression approaches.

\end{abstract}

\begin{document}

\maketitle
% \ba{suggestions: "Multi-modal transfer learning between biological foundation models"}\\
% \ba{"IsoFormer: A multi-modal transfer between biological foundation models"}\\
% \ba{"IsoFormer: A multi-modal approach to connect biological foundation models"}
\section{Introduction}

% Foundations models revolutionized many fields, and recently many fields across biology
% There are foundation models that have been developed separately for \dna, RNA and proteins with impressive results in each field
% These models are either pre-train with self-supervised learning (e.g. MLM methods, ESM, NT, …) or with supervised learning on large amounts of data (AlphaFold, Enformer, …)

Foundation models have ignited a revolution in numerous scientific fields, starting in \textit{NLP} and computer vision, and more recently in several domains within the life sciences. Within the biological sciences, these models have enabled predicting protein structures from sequences \cite{jumper2021highly, lin2023evolutionary}, deciphering the genome functions \cite{Dalla-Torre2023.01.11.523679, zhou2023dnabert, nguyen2024hyenadna, alphamissense} and interactions of biomolecules \cite{ross2022selective}, and crafting new molecules not found in nature \cite{merchant2023scaling}. Specifically, significant progress has been made with foundation models tailored to \dna \cite{Dalla-Torre2023.01.11.523679, zhou2023dnabert, nguyen2024hyenadna, Nguyen2024_evo}, \rna \cite{SpliceBERT, BigRNA, codonbert, shulgina2024rna}, and protein \cite{lin2023evolutionary} sequences. These foundation models have typically been developed and trained separately, using either self-supervised learning techniques such as Masked Language Modeling (\textit{MLM}), as seen in models like \esm \cite{lin2023evolutionary} and the Nucleotide Transformer (\nt) \cite{Dalla-Torre2023.01.11.523679}, or supervised learning approaches on large datasets, as in AlphaFold \cite{jumper2021highly} and \enformer \cite{avsec2021effective}. These models have been instrumental in advancing our understanding of biology by accurately predicting the structures and functions of biological sequences.

% In biology, the central dogma connects \dna, RNA and proteins
% This motivates an architecture that connects the three modalities to build a unified foundation model for biology to understand how cells work
% For the sake of efficiency and performance, our architecture should enable transfer from the different modalities to leverage the mono-modalities pre-trainings that have already been performed
While existing methods provide relevant insights, they are still limited by the fact that they only consider a single sequence modality. In biology, the \emph{central dogma} describes the flow of genetic information from \dna to \rna to proteins \cite{crick_central_dogma}. This fundamental concept underscores the interconnectedness of these three types of biological sequences and highlights the potential for a unified modeling approach. An architecture that integrates \dnanospace, \rnanospace, and protein modalities should provide a comprehensive model for biology that mirrors the natural processes within cells. Furthermore, by enabling transfer learning across modalities, the model can capitalize on the vast amounts of pre-training already performed on individual \dnanospace, \rnanospace, and protein datasets.
% This multi-modal approach not only aligns with the central dogma but also promises to deepen our understanding of cellular mechanisms and enhance predictive capabilities across biological sequences.

% chalenges of multi-modal models
Developing deep learning models using multiple biological sequence modalities has been mainly limited by the lack of matched available data; existing databases usually isolate a specific modality and thus relationships between modalities are not easily obtainable. As more multi-modal datasets are made available \cite{tomczak2015review, samaras2020proteomicsdb}, it is becoming possible to develop models that extract and combine the information from \dnanospace, \rnanospace, and protein sequences to better model the different cellular processes. Such multi-modal models have already been successful in other domains, such as mixing language and visual inputs \cite{radford2021learning, alayrac2022flamingo, liu2024visual, srivastava2024omnivec, tewel2022zerocap, lu2023multi, plip_pathology, conch_pathology}, but until now there are no models that can handle multiple biological sequence modalities.

\begin{figure}[h]
\centering
\includegraphics[width = 0.85\textwidth]{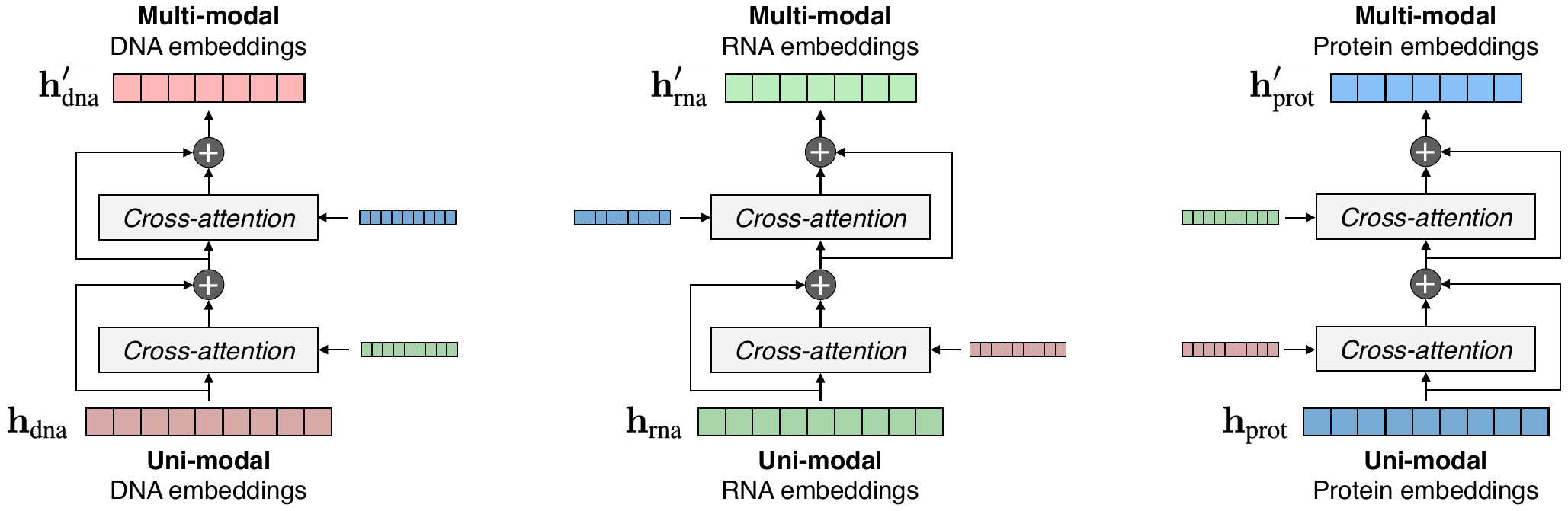}
\caption{\small Our aggregation module compiles information from the different biological sequence modalities of \dnanospace, \rnanospace, and proteins by using successive cross-attention layers and residual connections.}
\label{fig:aggregation_abstraction}
% \vspace{-15pt}
\end{figure}

% What we do
In our work, we propose the first multi-modal architecture to connect \dnanospace, \rnanospace, and proteins (Fig. \ref{fig:aggregation_abstraction}). Our approach is based on three main components: (i) pre-trained modality-specific encoders that produce one embedding per modality, (ii) aggregation layers that combine information from the encoders and create a multi-modal representation, and (iii) a task-specific head that predicts the desired output. We show that our multi-modal approach transfers and aggregates knowledge of pre-trained mono-modal encoders and outperforms previous single-modality baselines (\enformer \cite{avsec2021effective}, \nt \cite{Dalla-Torre2023.01.11.523679}, and \esm \cite{lin2023evolutionary}). We also demonstrate the flexibility of our approach by comparing different encoders for a specific modality and different aggregation techniques. While some previous approaches have modeled specific interactions between modalities, such as protein-to-DNA interaction using structure information and modules \cite{wei2022protein}, our approach is general-purpose and can be adapted to any task involving one or more biological sequence types.

% motivate isoform prediction task
Significant problems in genomics intrinsically involve multiple sequence modalities \cite{hasin2017multi, wei2022protein}, and it is still unclear how to adapt general-purpose sequence models to those cases. In order to validate our multi-modal approach, we focused on the study of a crucial task in genomics that has been challenging to tackle using a single sequence modality - namely the prediction of \rna\emph{transcript isoform expression} across different tissues \cite{mishra2020computational}. When gene \dna sequences are transcribed into \mrna molecules to produce proteins, they generally do not express it in just a single way producing a single protein isoform. After \dna sequences are transcribed, \premrna transcripts undergo a process called \rna alternative splicing where they are cut and re-assembled to form different variants of mature \mrna molecules that can be translated into proteins. This process allows for the generation of multiple \rna and protein isoforms from a single gene \dna sequence, that can differ in structure, function, localization, or other properties (Fig.~\ref{fig:main_figure}a). Therefore, predicting which isoforms are expressed in a given cell or under specific conditions is key to understand gene regulation and disease mechanisms. This task is multi-modal in nature, since only looking at the \dna sequence present in the cell does not provide a complete picture of the different \rna isoform landscape. Focusing on this single important task in opposition to tackling more but less challenging tasks should provide a stronger evidence for the effectiveness of our approach to advance biological knowledge.

In summary, our contributions are the following: (i)~We built the first multi-modal model for the integration of \dnanospace, \rnanospace, and protein sequences. (ii)~We show that our model achieves efficient transfer learning between the three modalities, not only leveraging intra-modalities pre-training but also inter-modalities transfer. (iii)~We use our architecture to tackle a new central task in biology that requires a multi-modal approach, namely \rna\emph{transcript isoform expression prediction}, and obtain state-of-the-art results, overcoming limitations of existing gene expression models such as \enformernospace. (iv)~Finally, we performed ablation studies to validate our different architectural choices and release our trained isoform expression prediction model \modelnamenospace, providing a new framework and baseline to the community and opening the door to future research on multi-modal sequence modeling and multi-modal biological problems.

\section{Related Work}

\textbf{Biological sequence modeling}\quad Researchers have explored different ways to process \dnanospace, \rnanospace, and protein sequences for multiple applications. Initial approaches included dynamic programming \cite{li2023protein}, hidden Markov models \cite{yoon2009hidden}, and genomic hash tables \cite{wu2016gmap}. Recently, deep learning, via supervised \cite{jumper2021highly} and unsupervised frameworks \cite{Dalla-Torre2023.01.11.523679,zhou2023dnabert,shulgina2024rna,lin2023evolutionary}, has gained thrust in the community. These methods, influenced by advancements in Large Language Models \cite{devlin2018bert,brown2020language}, have focused on \dna \cite{Dalla-Torre2023.01.11.523679, zhou2023dnabert, nguyen2024hyenadna, Nguyen2024_evo}, \rna \cite{SpliceBERT, BigRNA, codonbert, shulgina2024rna}, and protein sequences \cite{lin2023evolutionary}. They capture complex biological patterns and perform tasks like protein structure prediction \cite{lin2023evolutionary} and variant effect prediction \cite{brandes2023genome}. These efforts have become more targeted given the different challenges of each sequence modality. For instance, models like HyenaDNA \cite{nguyen2024hyenadna, Nguyen2024_evo} or Caduceus \cite{schiff2024caduceus} tackled the long-range dependencies in \dnanospace. Few models consider multiple modalities, mainly focusing on structural information (e.g. predicting \dna interacting residues in proteins \cite{wei2022protein}). Our work, \modelnamenospace, is the first general-purpose model integrating three biological sequence modalities.

% For years, practitioners focused on developing custom models for specific applications, using a supervised learning framework. Using the abundance of data and advancements in Large Language Models, unsupervised learning changed the paradigm of biological sequences modeling with protein LMs (pLMs) such as ESM \cite{lin2023evolutionary}, DNA LMs (dLMs) \cite{Dalla-Torre2023.01.11.523679, zhou2023dnabert} or more recently RNA LMs (rLMs) \cite{shulgina2024rna}. These models are able to capture complex patterns within biological sequences and perform a variety of downstream tasks such as protein structure prediction \cite{lin2023evolutionary} or variant effect prediction \cite{brandes2023genome}.

% Most efforts have tried to tackle specific challenges of each of the omics. For instance, HyenaDna or Caduceus aimed to tackle the long-range dependencies of DNA sequences. Many models have tried to integrate additional information such as species \cite{karollus2024species}. Few models have studied the interaction between omics, mainly focusing on structure information [A deep learning-based method for the prediction of DNA interacting residues in a protein]. However, \modelname is the first general-purpose model that proposes to integrate the three omics under the form of sequences only. 

\textbf{Multi-Modal Integration}\quad Efforts to address multi-modality in deep learning models have been prominent in \textit{NLP} \cite{DBLP:journals/corr/abs-2308-12966,DBLP:journals/corr/abs-2305-18565,DBLP:conf/icml/DriessXSLCIWTVY23,DBLP:conf/nips/Huang0WHSML0MPL23,DBLP:conf/icml/0008LSH23} and computer vision \cite{DBLP:conf/cvpr/LeiLZGBB021,DBLP:conf/nips/Yu0YB23,DBLP:journals/corr/abs-2212-04979,DBLP:journals/tmlr/YuWVYSW22,DBLP:journals/corr/abs-2212-03191}. In computer vision, integrating image and audio for video classification \cite{DBLP:conf/euvip/ShaikhCIA23} or audio-visual segmentation \cite{DBLP:conf/aaai/GaoCCWL24} is common, often using cross-attention \cite{DBLP:conf/aaai/WangLLD0L24}. Unsupervised approaches using contrastive learning also integrate image, audio, and other modalities \cite{DBLP:conf/icip/LiuTL23}. Large Language Models have driven multi-modal efforts for text and image with methods like Flamingo \cite{alayrac2022flamingo} and \textit{CLIP} \cite{radford2021learning}, leading to various integration architectures, including the Perceiver Resampler \cite{jaegle2021perceiver,alayrac2022flamingo} and C-Abstractor \cite{cha2023honeybee,mckinzie2024mm1}. \modelname integrates \dnanospace, \rnanospace, and protein sequences, leveraging past integration architectures.

\textbf{Gene Expression Prediction}\quad Transcript isoform expression prediction is a more refined and challenging task within gene expression prediction. Traditionally, gene expression has been addressed through tailored approaches and experimental annotations \cite{cheng2012understanding, gonzalez2015early}. Recently, deep learning models have demonstrated improved results by predicting gene expression directly from \dna sequence \cite{kelley2018sequential, zhou2018deep}. \enformer \cite{avsec2021effective} leveraged dilated convolutions and attention layers to extend the context window to 190 kilo base pairs (kbp), achieving new state-of-the-art results on gene expression. However, these models are limited to gene-level expression levels and cannot predict isoform-specific expression as \dna sequence information is not sufficient to solve this task. We introduce \modelnamenospace, the first multi-modal model that is able to predict transcript isoform expression.

\section{Background}

\textbf{Central dogma of biology}\quad In this work, we consider three biological sequence types: \dnanospace, \rnanospace, and proteins. These sequences (composed of nucleotides or amino-acids) are fundamental to biological processes in living organisms. They are also intricately intertwined: \dna dictates \rna synthesis during \emph{transcription}, and \rna guides protein synthesis through \emph{translation} (see Fig.~\ref{fig:main_figure}; \cite{crick_central_dogma}). Thus, changes in \dna can alter \rna and protein sequences, impacting an organism's phenotype (i.e. function). \dna and \rna sequences consist of four nucleotides (\textit{ACGT} and \textit{ACGU}, respectively), while proteins are made of 20 amino-acids. Thus, these sequences are commonly modelled as linear strings. Obtaining data in this format is becoming more available thanks to recent advances in high-throughput sequencing technologies.

\textbf{Gene expression and isoforms}\quad Gene expression is the process by which a gene's \dna sequence is transcribed into various \rna molecules that code for specific proteins. This process is complex, as a single gene can produce different \rna and protein \emph{isoforms} with varying abundances across tissues (Fig.~\ref{fig:main_figure}a). \rna isoforms are \mrna molecules of different exon compositions derived from the same gene, produced via processes like alternative splicing. The expression level of each isoform is commonly measured by counting the amount of its \rna molecules in cells. Accurate isoform expression prediction across tissues can aid in understanding genetic variants' effects on cellular processes and phenotypes. However, this task cannot be tackled solely using \dna sequences as the same \dna sequence produces different isoforms in different cellular contexts and types. In addition, both \dna and \rna sequence features are crucial as \dna sequences contain regulatory elements that control transcription levels, while \rna isoform sequences have features affecting their stability and degradation.

\section{Method}
\subsection{Multi-modal framework}
\label{sec:mutimodal_framework}

We consider respectively \dnanospace, \rnanospace, and protein sequences $\xdna \in \dnaalphabet$, $\xrna \in \rnaalphabet$ and $\xprot \in \protalphabet$ where $\dnaalphabet$ is the \dna base alphabet \textit{ACGT}, $\rnaalphabet$ is the \rna base alphabet \textit{ACGU}, and $\protalphabet$ is the set of 20 amino-acids. We then consider three associated modality-encoders $f$ with respective weights $\left(\thetadna, \thetarna, \thetaprot \right)$ that encode sequences $\mathbf{x}$ into corresponding embeddings $\mathbf{h}$:
\begin{equation}
    \hdna = f_{\thetadna}\left(\xdna\right), \ \ \hrna = f_{\thetarna}\left(\xrna\right), \ \ \hprot = f_{\thetaprot}\left(\xprot\right)
\end{equation}
We assume that the weights $\left(\thetadna, \thetarna, \thetaprot \right)$ have been obtained through independent pre-training processes that can involve supervised or self-supervised training techniques over large corpus of biological data. Note that in practice, as commonly used in recent works \cite{richard2024chatnt}, models pre-trained over \dna sequences can be re-used to produce embeddings for \rna sequences, replacing artificially the uracil base (\textit{U}) by thymine (\textit{T}) in the input. In this work, we aim to connect these encoders and train them jointly to learn a multi-modal embedding $\hmm$. We start by learning a multi-modal embedding per modality defined as
\begin{equation}
    \hdna^{\prime} = f_{\phi}\left(\hdna, \hrna, \hprot \right),\ \ \hrna^{\prime} = f_{\phi}\left(\hrna, \hdna, \hprot \right),\ \ \hprot^{\prime} = f_{\phi}\left(\hprot, \hdna, \hrna \right)
\end{equation}
where $f_{\phi}$ is an aggregation function with weights $\phi$. Then, we define the multi-modal embedding as the concatenation of the per modality multi-modal embeddings:
\begin{equation}
    \hmm = \left[ \hdna^{\prime}, \hrna^{\prime}, \hprot^{\prime} \right].
\end{equation}
Note that this definition is general as it allows the use of any aggregation function over the modality embeddings $\mathbf{h}$. The multi-modality embeddings per modality $\mathbf{h^{\prime}}$ are introduced to solve tasks that are "modality-centered". For instance, $\hdna^{\prime}$ could be used to solve a task that involves nucleotide-level annotation over a \dna sequence while requiring other modalities as input. We rely otherwise on the concatenated multi-modal embedding $\hmm$ to solve any other task.

\subsection{Genes and isoforms expression}

\begin{figure}[h!]
  \centering
  \includegraphics[width=1\textwidth]{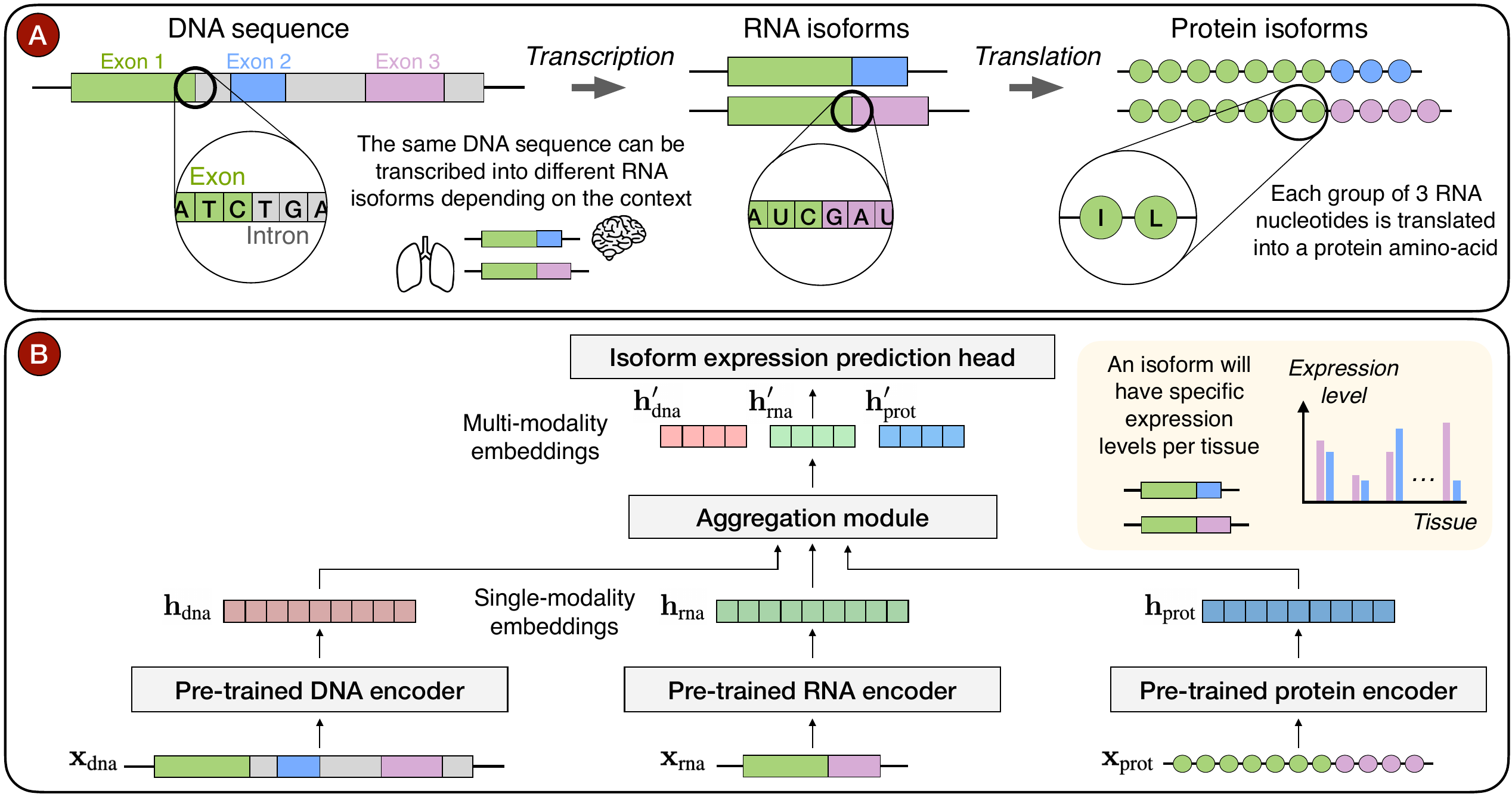}
  \caption{\small \textbf{a)} Three types of biological sequences are considered in this work: \dnanospace, \rnanospace, and proteins. These sequences are composed of nucleotides (\dna and \rnanospace) or amino-acids (protein). In a single gene, several coding regions or exons can be used to create different \rna transcript isoforms and proteins. The abundance of each isoform is tissue-dependent and its measurement is called expression level. \textbf{b)} \modelname leverages pre-trained encoders that produce modality-specific embeddings, which are then aggregated into multi-modal embeddings. These are used to predict the expression of a given \rna transcript isoform across multiple tissues.}
  \label{fig:main_figure}
  % \vspace{-15 pt}
\end{figure}

We now introduce our notations specific to the task of \rna isoform expression prediction (Fig.~\ref{fig:main_figure}a). We consider a \dna sequence $\xdna$ of length $L$ to contain a gene $g$. In practice, given the input size limitation of the existing foundation models, the sequence length $L$ might be shorter than the full length of the gene. In this case, we choose the \dna sequence $\xdna$ to be centered on the start of the gene, i.e., where transcription begins, which also surrounds the promoter regions known to be important for transcription. This way, we also capture all the enhancer regulatory elements upstream of the transcription site. 

We denote by $\xrna^{(1)}$, ..., $\xrna^{(n)}$ the $n$ existing transcripts for gene $g$ across all tissues. Coding transcripts are translated into proteins and we denote by $\xproti$ the amino-acid sequence of the protein associated to the transcript $\xrnai$. We define the expression $e$ of genes and transcripts across tissues $T$ as:
\begin{equation}
   \forall T, \expr{\xdna, T} = \sum\limits_{i=1}^n \expr{\xrnai, T} \in \mathbb{R}.
\end{equation}
While deep learning models have been trained to predict the overall expression of genes across tissues with great accuracy \cite{kelley2018sequential, zhou2018deep, avsec2021effective, linder2023predicting}, to our knowledge no model can predict the expression of the different \rna transcripts across tissues directly from the sequence. In this work, we leverage our multi-modal framework to train a transcript expression level prediction model, dubbed \modelnamenospace, that takes as input a \dna sequence, an \rna transcript sequence, and its matching protein sequence to predict the expression of that transcript across tissues as measured by bulk \rnaseq (Fig.~\ref{fig:main_figure}).

\textbf{Aggregation module}\quad To capture the local patterns and their relationship across modalities, we introduce an embedding aggregation method based on cross-attention with residual connections (Fig.~\ref{fig:aggregation_abstraction}). The module is applied to each modality, performing cross-attention successively to the other modalities to produce multi-modal embeddings that are specific to each modality by keeping their dimensionality and which can be also stacked. Note that this aggregation method is robust to the absence of a specific modality as the cross-attention term will be zeroed out. The final multi-modal embedding $\hmm$ can then be used to solve any task. Our model can be trained end-to-end and it could be applied to different tasks by changing the prediction head.

% For tasks where the output size is independent from the input dimension such as expression prediction, we use a simple global average pooling over the length of the embedding and a linear head that outputs one value per tissue. 

\subsection{Transferring from pre-trained biological encoders}

\begin{table}[h!]
\caption{\small Characteristics of single-modality encoders considered in \modelnamenospace. $\dagger$ This column indicates the tokenization scheme (e.g., 1 token corresponds to 6 nucleotides).}
  \label{tab:mono_omic_encoders}
  \vspace{0.1cm}
  \centering
  \setlength{\tabcolsep}{3pt} % Reduces the space between columns
  { \small
  \begin{tabular}{cccccc}
    \hline
    \textbf{Modality}    & \textbf{Model} & \textbf{Pre-training} &\textbf{Num. Params.} & \textbf{Tokens}$^\dagger$ & \textbf{Sequence Length} \\
    \hline
   \dna & \enformer \cite{avsec2021effective} & Supervised & 110M & 1 Nucleotide & 190,000 \\
   \dna & \nt (v2) \cite{Dalla-Torre2023.01.11.523679} & Self-Supervised & 250M & 6 Nucleotides & 12,282 \\
    \rna & \nt (v2) \cite{Dalla-Torre2023.01.11.523679} & Self-Supervised & 250M & 6 Nucleotides & 12,282 \\
    Protein & \textit{ESM2} \cite{lin2023evolutionary} & Self-Supervised & 150M & 1 Amino-acid& 2,047 \\
    \hline
  \end{tabular}
  }
  \vspace{2pt}
\end{table}

\textbf{Encoders}\quad We describe here the encoders used to process each sequence modality (see also Table~\ref{tab:mono_omic_encoders} and Fig.~\ref{fig:main_figure}b). For protein sequences, we used \esm (\textit{ESM2-150m}) \cite{lin2023evolutionary}, which handles any protein up to 2,048 amino-acids. The \esm models have been pre-trained through \textit{MLM} on large corpuses of protein sequences and are considered state-of-the-art on multiple tasks including folding. For \dna sequences, we used the \nt model \cite{Dalla-Torre2023.01.11.523679}, which has also been trained through self-supervision to reconstruct masked 6-mers within 12 kbp genomic regions from 850 species. Additionally to the \nt, we also considered the \enformer model as \dna encoder. The \enformer is a different type of model that has been pre-trained through full supervision to predict multiple experiments related to chromatin accessibility and modifications, transcription factor binding, and more importantly gene expression. \enformer is a strong candidate for our architecture's \dna encoder module as it can process sequences up to 200kbp as well as one can expect to obtain transfer from its gene expression capabilities. Finally, while foundation models have been reported to be pre-trained on \rna sequences, none of them has been made publicly available at the moment and pre-training a foundation model from scratch is out of the scope of this study. As such, we re-used the \nt model \cite{Dalla-Torre2023.01.11.523679} to compute embeddings for \rna sequences as this model has been reported recently to also be able to solve \rna \cite{richard2024chatnt} and protein \cite{boshar2024genomic} tasks with simple adaptations, using the trick described in section~\ref{sec:mutimodal_framework}.

\textbf{Architecture}\quad Our architecture leverages three pre-trained encoders, one per bio-modality, as well as the aggregation function defined above, to generate a multi-modality embedding $\hmm$ (Fig.~\ref{fig:main_figure}b). That embedding is finally transformed by an expression head $f_{\psi}$ with weights $\psi$ to make isoform expression level predictions across tissues. As the shape of the network output must be independent from the dimension of its inputs, we used a global average pooling over the length of the embedding. A linear layer then outputs one value per tissue. As long as the encoders can take in a biological sequence to produce an embedding, our method can function with different types of general-purpose encoders. While this flexibility allows us to leverage a big part of the landscape of biological sequence encoders, for this work we chose the specific models described in the section below.

\subsection{Training}

\textbf{Objective}\quad We denote the \modelname model by $f_{\theta, \phi, \psi}$ where $\theta$ is the concatenation of the weights of the encoder models. These weights are initialized to the values obtained after pre-training of the different encoders. Respectively $\phi$ and $\psi$ denote the weights of the aggregation module and expression prediction head. The \modelname is trained to minimize the following objective
\begin{equation}
  \mathcal{L_{\text{MSE}}} = \sum\limits_{T} \left(f_{\psi}\left(\hmm, T \right) - \expr{\xrnai, T}\right)^2,\text{where} \ \ \hmm = f_{\theta, \phi}\left(\xdna, \xrnai, \xproti\right)
\end{equation}
where the summation is performed over a set of available tissues. Note that this framework can also accept only one or two of the three modalities as input. This is the case for instance when predicting expression of non-coding transcripts that do not translate into proteins.

\textbf{Dataset}\quad We conducted our analysis of \modelname on \rna transcript expression data obtained from the GTEx\footnote{\url{https://www.gtexportal.org/home/downloads/adult-gtex/bulk_tissue_expression}} portal. We used Transcript \textit{TPM}s measurements across $30$ tissues, which come from more than 5,000 individuals. We followed a common process in gene expression datasets \cite{zhou2018deep}: we averaged the expression levels for a given tissue across individuals, and used the reference genome sequence as input. We mapped transcripts to their original genes and associated proteins using the Ensembl database \cite{martin2023ensembl}. Our resulting dataset is made of triplets of \rna transcript sequences, \dna sequences (centered on the Transcription Start Site (\textit{TSS}) of the transcript), and proteins. In total, the dataset is made of $\sim$170k unique transcripts, of which 90k are protein-coding and correspond to $\sim$20k unique genes. Our dataset has a fixed train and test set, divided by genes; all presented results correspond to the performance on the test set. We provide more details on the dataset in Appendix~\ref{appendix:dataset}.

\textbf{Hyperparameters}\quad We used the Adam optimizer with a learning rate of $3\cdot10^{-5}$ and batch size of $64$, and used early stopping on a validation set comprised of $5\%$ of the train set to reduce training time. We also made our baseline model's weights available\footnote{\url{https://huggingface.co/isoformer-anonymous/Isoformer}}. More details in Appendix \ref{appendix:training}.

% \modelname can be trained end-to-end on specific tasks or in a multi-task fashion. Also \modelname can handle missing modalities within samples as all our aggregation techniques are robust to missing modality as they are zeroed-out in aggregation layers. We keep flexibility around whether to fine-tune the encoders or only aggregation layers, with a trade-off between performance and computational costs.

% \subsection{Training}

\section{Experiments}

We present extensive experiments to assess the performance of our multi-modal approach on the \textit{transcript isoform expression prediction} task and compare it with existing single-modality approaches. We show that (i) our architecture efficiently aggregates modalities to improve its performance on this task; (ii) by using a tailored model for expression prediction as a base \dna encoder, \modelname reaches state-of-the-art performance; (iii) we provide an extensive ablation study on different aggregation approaches; (iv) we demonstrate that our approach achieves transfer learning both intra-modalities from their independent pre-training as well as inter-modalities.   

\subsection{Bridging three foundational models outperforms mono-modal approaches}

\textbf{Experiment}\quad We investigated the effect of adding the different modalities within our multi-modal framework for the prediction of expression of each \rna transcript across different human tissues. We used \nt as the foundation model for both \dna and \rna and \esm as the protein encoder. We compared our multi-modal approach (\dna + \rna + protein) with models trained with different combinations of modalities as input: \dna only, \rna only, protein only, \dna + protein and \dna + \rnanospace. Results were obtained over 5 random seeds; for each random seed we change the validation set and randomly initialize the non pre-trained parameters $\left(\phi, \psi\right)$ of our model. We report both $R^2$, which measures how well each model predicts the actual values of expression, and Spearman correlation across tissues, which is a metric for ranking transcripts based on their expression in each tissue.

\begin{table}[h!]
    \caption{\small Performance of different variants of \modelname for the prediction of transcript isoform expression. $R^2$ and Spearman correlation across tissues for 5 different random seeds is reported. \nt is used as both \dna and \rna encoder while \esm is used to process protein sequences.}
      \label{tab:omics_integration}
  \vspace{0.1cm}
  \centering
  { \small
  \begin{tabular}{l|cc}
    \hline
    \textbf{Model Input}    & $\mathbf{R^2}$ &\textbf{Spearman} \\
    \hline
    \dna only & $0.13 \pm 0.02$ & $0.43 \pm 0.01$  \\
    \rna only   & $0.36 \pm 0.03$  & $0.61 \pm 0.01$ \\
    Protein only & $0.20 \pm 0.01$  & $0.46 \pm 0.01$  \\
    \hline
    \dna + Protein & $0.28 \pm 0.01$  & $0.52 \pm 0.01$ \\
    \dna + \rna & $0.39 \pm 0.01$ &  $0.64 \pm 0.01$ \\
    \dna + \rna + Protein & $\mathbf{0.43 \pm 0.01}$  & $\mathbf{0.65 \pm 0.01}$\\
    \hline
  \end{tabular}
  }
  \vspace{2pt}
\end{table}

\textbf{Results}\quad We observed that our approach benefits from adding more modalities as the performance increases from one modality alone to having two combined, and the best performance is achieved with the three (\dnanospace, \rnanospace, and protein) modalities together (Table~\ref{tab:omics_integration}). This is true for both Spearman correlation and $R^2$ metrics, with stronger improvement for the latter reflecting a more accurate prediction of the actual values of expression and not just the ranking of transcripts. This is a strong demonstration that our model can aggregate information across modalities to improve performance on this isoform expression task. In addition, we observe increased performance by using \dna together with \rna compared with \dna and protein information. This can be related to the strong importance of the \textit{UTR} regions of the \rna sequence in the regulation of its degradation and stability \cite{koh2019tuning}, which affect its final expression level in the cells, that are not captured at the protein level.

\subsection{Enformer as \dna encoder module to obtain transfer between expression prediction tasks}
\label{sec:results_transfer}

\textbf{Experiment}\quad To showcase the flexibility of \modelname towards different modality-specific encoders, we tested replacing \nt by the \enformer model \cite{avsec2021effective} as \dna encoder. \enformer has been trained over gene-level expression data obtained from \textit{CAGE} assays (one value of expression per gene per tissue), while our model is trained to predict \rna transcript expression data obtained from bulk \rnaseq assay (one value of expression for each isoform of a given gene per tissue) and therefore represents a different challenge that cannot be tackled from the \dna sequence alone. Still, as the \enformer has been trained to predict gene-level expression across tissues directly from \dna sequences, a related task to predicting \rna isoform expression, one might expect to obtain transfer by using it as \dna encoder.

\begin{table}[h!]
    \caption{\small Comparison of \enformer and \nt \dna encoders used in \modelnamenospace. $R^2$ and Spearman correlation across tissues on the transcript isoform expression prediction task. Standard deviation across 5 seeds is reported.}
      \label{tab:nt_vs_enformer}
  \vspace{0.1cm}
  \centering
  { \small
  \begin{tabular}{l|cc}
    \hline
    \textbf{Model}    & $\mathbf{R^2}$ &\textbf{Spearman} \\
    \hline
    \enformer & $0.21 \pm 0.01$  & $0.46 \pm 0.00$\\
    \hline
    \modelname (NT) & $0.43 \pm 0.01$  & $0.65 \pm 0.01$ \\
    \modelname (Enformer) &  $\mathbf{0.53 \pm 0.01}$ &  $\mathbf{0.72 \pm 0.00}$ \\
    \hline
  \end{tabular}
  }
  \vspace{2pt}
\end{table}

\textbf{Results}\quad We obtained superior performance using the \enformer instead of \nt as a pre-trained \dna encoder both as \dna-only but also when we combined with the \rna and protein encoders (Table~\ref{tab:nt_vs_enformer}). Importantly, also with the \enformer our framework benefits from bridging modalities. This improved performance can be explained by the fact that \enformer has been pre-trained on the related task of gene expression prediction and thus its embeddings are better aligned with the isoform prediction task. Moreover, the \enformer is a model that can handle sequences of large context (up to 196k nucleotides), enabling it to capture long-range dependencies, known to be relevant for expression. These results demonstrate that our multi-modal framework can be improved by leveraging more domain-specific encoders. As our best model for isoform prediction is achieved using the \enformer as \dna encoder, we will use it as \dna encoder by default for all the following experiments. We make the weights of this \modelname model available on HuggingFace\footnote{\url{https://huggingface.co/InstaDeepAI/isoformer}.}.

\begin{figure}[t!]
    \centering
    \begin{minipage}[b]{0.33\textwidth}
        \centering
        \vspace*{\fill}
        \includegraphics[width=\textwidth]{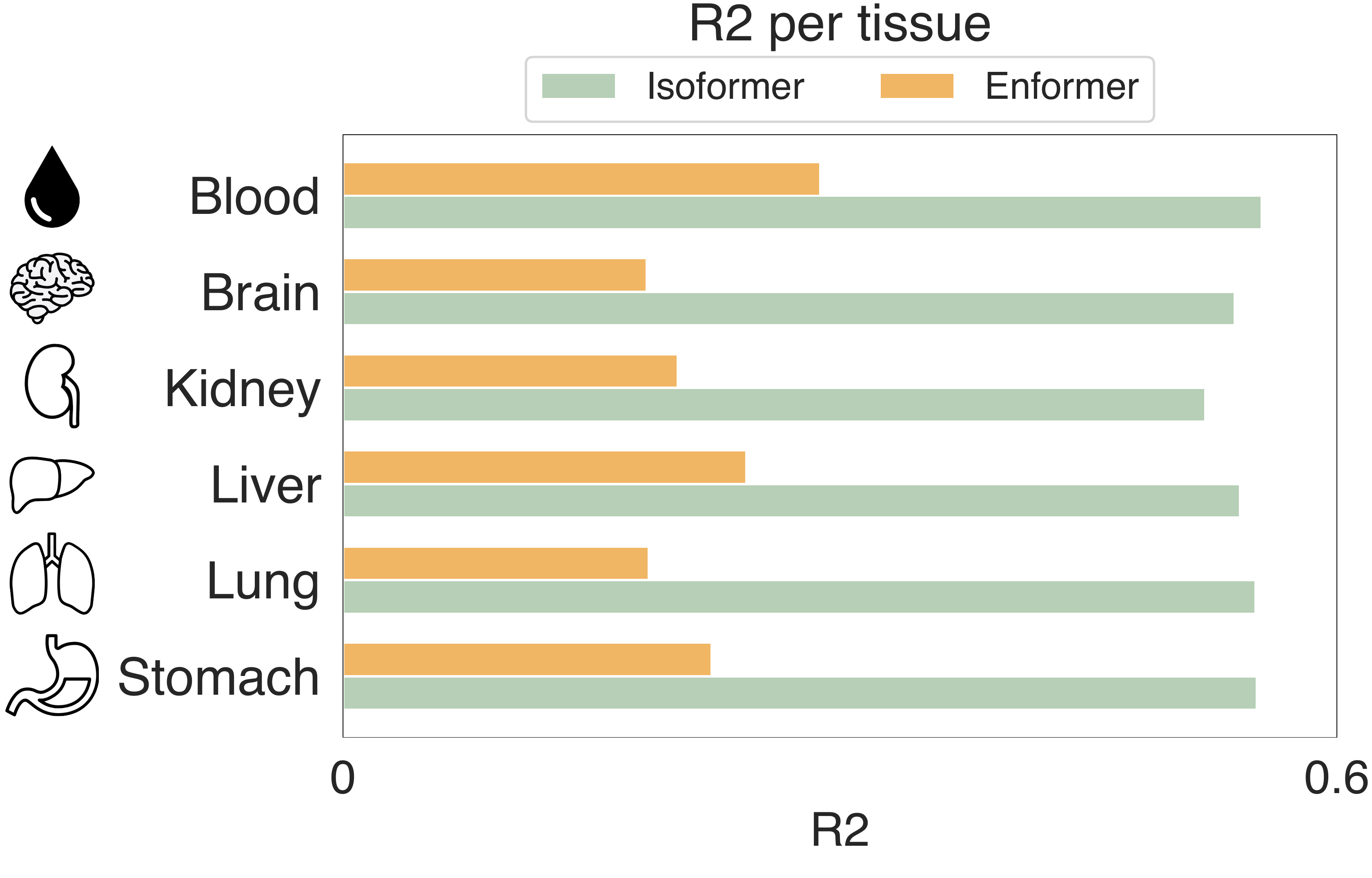}
        \vspace*{\fill}
        \label{fig:r2_per_tissue_subset}
    \end{minipage}
    \hspace{0.01\textwidth}
    \begin{minipage}[b]{0.63\textwidth}
        \centering
        \vspace*{\fill}
        \includegraphics[width=\textwidth]{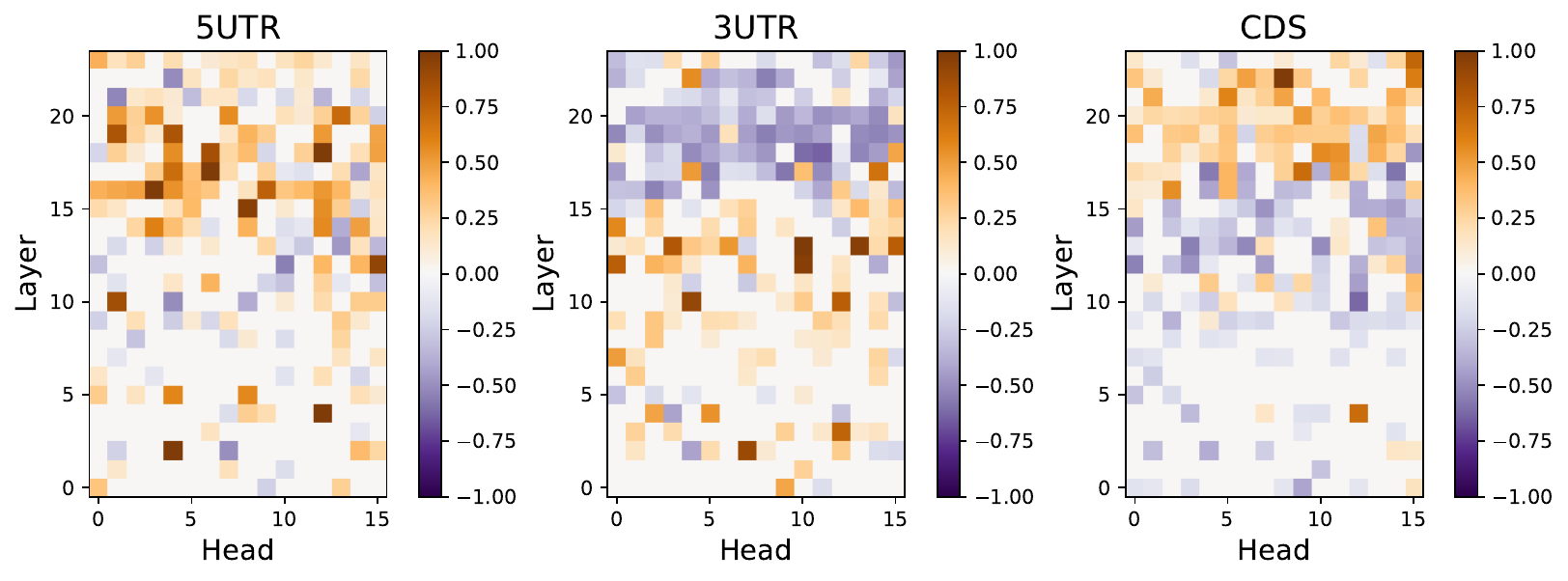}
        \vspace*{\fill}
        \label{fig:masked_comparison_rna}
    \end{minipage}
    \vspace{-10pt}
    \caption{\small \textbf{Left:} Performance of \modelname and Enformer~\cite{avsec2021effective} per tissue on a selected subset of tissues. \textbf{Right:} Changes in attention in the \rna encoder during fine-tuning. These scores are reported for three genomics elements of interest for all heads and layers of the \rna encoder.}
    \label{fig:combined}
\end{figure}

\textbf{Interpretation}\quad We report the performance of \modelname across selected tissues in Fig.~\ref{fig:combined}-left (results across all tissues are presented in Appendix \ref{appendix:additional_results}). \modelname obtains similar performance across tissues despite tissues having different distributions of expression levels. To gain additional insights about the representations learned by \modelnamenospace, we analysed the attention layers inside the \rna encoder as it is the one providing stronger improvement on this task. Specifically, we compared how the attention distribution within each layer and head changes when we finetune the \rna encoder alone versus finetuning \modelname altogether. We report changes in attention scores at each layer and head of the \rna encoder for three genomics elements known to have a strong effect on the isoform splicing and gene expression processes, namely the {\footnotesize 3}\textit{UTR}, {\footnotesize 5}\textit{UTR} and \textit{CDS} sequence, see Fig.~\ref{fig:combined}-right (additional details on these scores are in Appendix Section~\ref{sec:att_map_analysis}). The results show that, when finetuning using the three modalities, different layers specialize to capture specific features relative to isoform splicing and expression. Notably, the middle set of layers put higher attention weights to {\footnotesize 3}\textit{UTR} regions whereas the top layers of \nt attributes higher attention weights on \textit{CDS} and {\footnotesize 5}\textit{UTR}. We assume that this \rna encoder specialization during finetuning is key to achieve a strong representation towards the prediction of its tissue-specific expression.

\subsection{Ablation studies on the aggregation strategy} 

\begin{figure}[h]
\includegraphics[width = \textwidth]{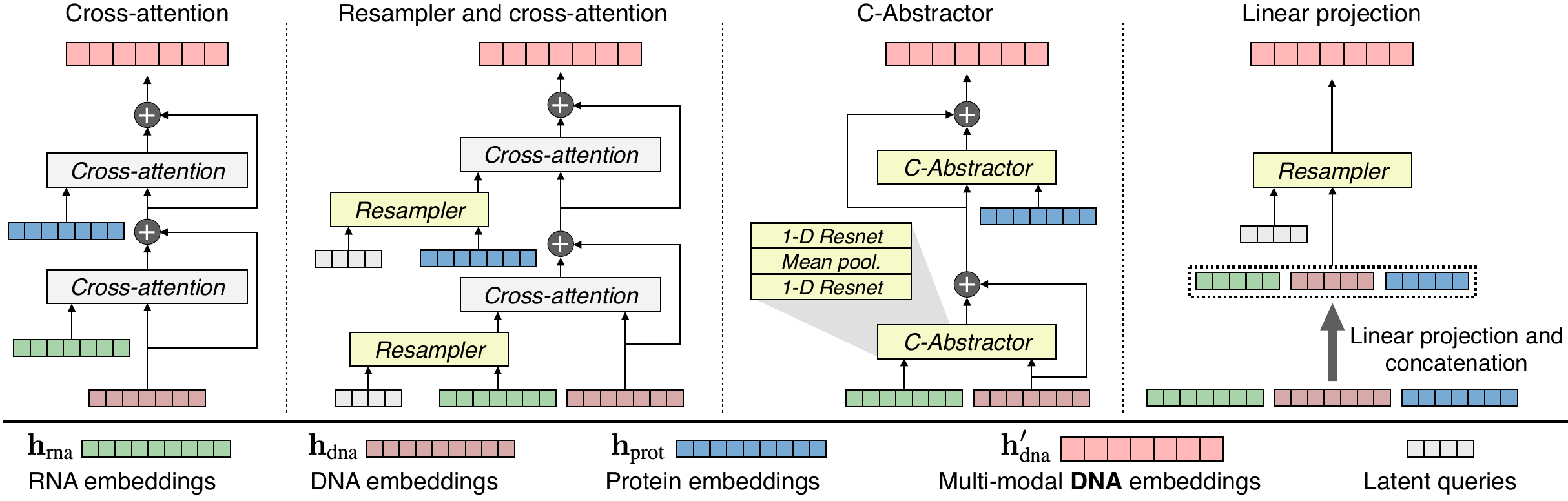}
\caption{\small Different aggregation strategies compared during the ablation studies. The figures show the specific case for obtaining multi-modal \dna embeddings $\hdna^{\prime}$; the same structure is used to obtain multi-modal \rna and protein embeddings ($\hrna^{\prime}$ and $\hprot^{\prime}$, respectively). In all cases, the \textit{Resampler} module is a \textit{Perceiver Resampler}~\cite{jaegle2021perceiver} and the \textit{Mean pooling} block is the \textit{Adaptive mean pooling} operator used in \cite{liu2024visual}.}
\label{fig:aggregation_techniques}
\vspace{-10pt}
\end{figure}

\textbf{Experiment}\quad We compared the \modelnamenospace's aggregation module with alternative strategies from recent multi-modal literature (Fig.~\ref{fig:aggregation_techniques}). Inspired by recent vision-text models \cite{liu2024visual, mckinzie2024mm1, peng2023kosmos, li2022blip}, we considered these three approaches:
(i) \textit{Perceiver Resampler} \cite{jaegle2021perceiver}: a variant of our cross-attention method using a Perceiver Resampler module. This block learns a fixed number of tokens for each modality, thus reducing the cost of the subsequent cross-attention layer. (ii) \textit{Linear Projection}: a strategy that linearly projects the embeddings of the three modalities into a common representation space and concatenates all tokens. This concatenated sequence is fed to a Perceiver Resampler to learn a fixed number of tokens which are then used in the head of the model. (iii) \textit{C-Abstractor} a 1-dimensional version of the C-abstractor architecture that provides a compromise between flexibility --choosing an arbitrary number of resampled tokens-- and locality preservation \cite{liu2024visual}.

\begin{table}[h!]
  \caption{\small Ablation study of different aggregation strategies when considering the three modalities together. All experiments were run using \enformer as the \dna encoder. PR = Perceiver Resampler.}
  \label{tab:aggregation_methods}
  \vspace{0.1cm}
  \centering
  { \small
  \begin{tabular}{l|cc}
      \hline
    \textbf{Aggregation Method} & $\mathbf{R^2}$ & \textbf{Spearman}
    \\
    \hline
    
    Ours  & $0.53 \pm 0.01$  & $0.72 \pm 0.00$ \\
    PR + Cross-Attn   & $0.49 \pm 0.04$ & $0.69 \pm 0.02$ \\
    Linear Proj. + PR & $0.49 \pm 0.02$  & $0.69 \pm 0.01$ \\
    C-Abstractor  & $0.53 \pm 0.01$  & $0.72 \pm 0.01$\\
    \hline
  \end{tabular}
  }
\end{table}

\textbf{Results}\quad We performed hyperparameters search with the same budget used for the \modelname aggregation module for all ablations methods. We report in Table~\ref{tab:aggregation_methods} the results obtained with the best set of hyperparameters for each method. We observe that, using an additional step of Perceiver Resampler always performs worse than only cross-attention --even after optimizing hyperparameters. Similarly, C-Abstractor does not confer any benefit over cross-attention; therefore we consider this latter strategy as the optimal aggregation strategy for our method. One advantage of the cross-attention mechanism we use is its interpretability, since it helps understanding which regions of the different modalities are being leveraged to make predictions. These conclusions align with recent multi-modal studies for other modalities \cite{DBLP:journals/corr/abs-2403-09611}.

\subsection{\modelname leverages knowledge of pre-trained encoders} 
 
\textbf{Experiment}\quad We showed in previous sections that \modelname is able to aggregate information from different biological sequence modalities to formulate high-quality predictions. Here, we investigated to which extent \modelnamenospace's performance can be attributed to the transfer from each modality-specific encoders' pre-training. We compared the performance of \modelname trained using all three encoders pre-trained on their respective domains with different \modelname model variants where (1) none of the encoders is pre-trained or (2) only the \dna or (3) \rna encoder is not pre-trained (i.e. randomly initialized). For this analysis we considered always the \enformer as the \dna encoder.

% Table~\ref{tab:pre_trained} compares the performance of using pre-trained and non-pre-trained encoders, considering both \enformer as the \dna encoder. 

\textbf{Results}\quad Our results demonstrate that using pre-trained encoders confers a substantial advantage to \modelnamenospace, as the $R^2$ is substantially larger (0.53) when compared with \modelname with none of the encoders pre-trained (0.10; Table~\ref{tab:pre_trained}). This demonstrates that \modelname is leveraging the knowledge acquired by each foundation model in the respective domains. However, we observed that when we randomly initialized only the \dna or \rna encoders, the drop in performance is smaller (0.41 for \dna and 0.48 for \rnanospace). This suggests that \modelname not only leverages intra-modalities pre-training but also inter-modalities transfer. Altogether, these results underpin our approach of relying on initializing \modelname with pre-trained encoders, as the information learned during pre-training is transferred and leveraged when considering multi-modal tasks.

\begin{table}[h!]
  \caption{\small Comparing the use of pre-trained and non-pre-trained encoders within \modelnamenospace. For this set of experiments the considered encoders are the Enformer for \dnanospace, \nt for \rna and \esm for proteins.}
  \label{tab:pre_trained}
  \vspace{0.1cm}
  \centering
  { \small
  \begin{tabular}{l|cc}
    \hline
    \textbf{Setting} & $\mathbf{R^2}$ & \textbf{Spearman} 
    \\
    \hline
    All models \textit{not} pre-trained & $0.10$   & $0.31$ \\
    \dna encoder \textit{not} pre-trained & 0.41 & 0.64 \\
   \rna encoder \textit{not} pre-trained & 0.48 &  0.69\\
    \hline
    All models pre-trained & $\mathbf{0.53}$  & $\mathbf{0.71}$  \\
    \hline
  \end{tabular}
  }
  \vspace{-10pt}
\end{table}

\section{Conclusion}

\modelname is the first model designed for multi-modal biological sequence modeling connecting \dnanospace, \rnanospace, and protein sequences. \modelname achieves state-of-the-art results by effectively leveraging and transferring knowledge from pre-trained \dnanospace-, \rnanospace-, and protein-specific encoders on one of the significant multi-modal problems in genomics: \rna\textit{transcript isoform expression prediction}. As part of our efforts, we are are open-sourcing our model, and hope \modelname paves the way to new milestones in building multi-modal models for biology.

\newpage

% \section*{References}
\bibliography{main}

%%%%%%%%%%%%%%%%%%%%%%%%%%%%%%%%%%%%%%%%%%%%%%%%%%%%%%%%%%%%
\newpage

\appendix

\section{Dataset details}
\label{appendix:dataset}

We based our dataset on the Genotype-Tissue Expression (GTEx) portal. Specifically, we use the 8th release of the \textit{Transcript TPMs} table\footnote{\url{https://www.gtexportal.org/home/downloads/adult-gtex/bulk_tissue_expression}}. This table is made of expression measurements of $170k$ transcripts across $30$ different tissues from $5,000$ individuals. As the goal is to build a general model for expression prediction from biological sequences, we averaged measurements across individuals to get an average expression value for each transcript in each tissue. 

We mapped each transcript to its corresponding gene and protein using the Ensembl\footnote{\url{https://www.ensembl.org/index.html}} database. Using this database, we were able to retrieve associated \dnanospace, \rnanospace, and protein sequences. For \dna sequences, we used the latest release of the human reference genome \textit{GRCh38}\footnote{\url{https://www.ncbi.nlm.nih.gov/datasets/genome/GCF_000001405.26/}}. 

To summarize, the steps to re-create our training dataset are the following:
\begin{enumerate}
    \item Download \textit{Transcript TPMs} table from GTEx portal\footnote{\url{https://www.gtexportal.org/home/downloads/adult-gtex/bulk_tissue_expression}}
    \item Compute average measurement per transcript ID and tissue
    \item Map \rna transcript isoform ID to its corresponding \rna sequence using Ensembl
    \item Get associated protein isoform using Ensembl
    \item Get chromosome and transcription start site position on the \dna sequence
\end{enumerate}

\begin{figure}[h]
\centering
\includegraphics[width = 1\textwidth]{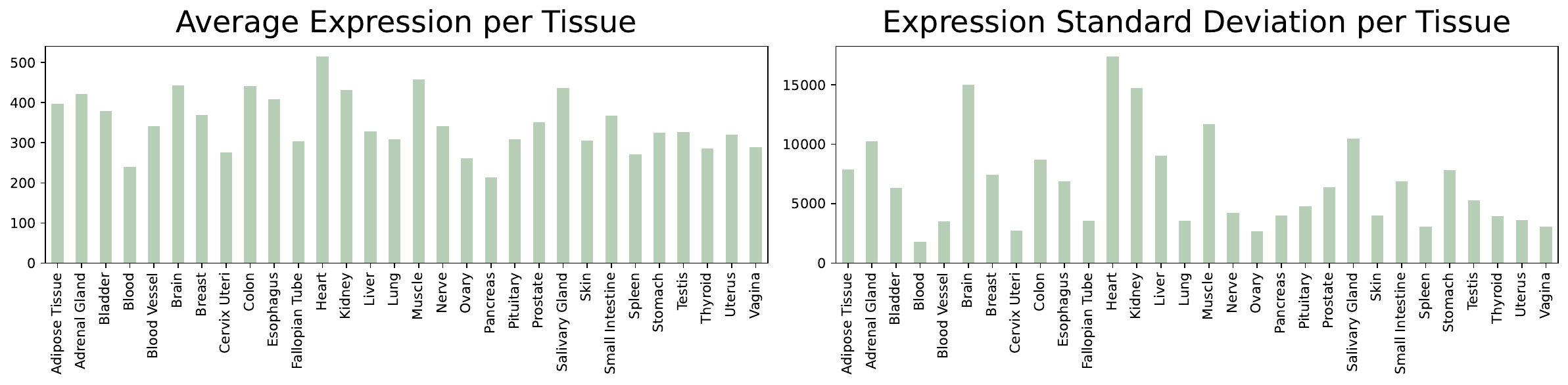}
\caption{\small Average and standard deviation of expression values across transcripts per tissue.}
\label{fig:expression_stats}
\end{figure}

\section{Training and architecture details}
\label{appendix:training}

All experiments were carried out with 5 seeds on 4 A100 GPUs (80GB RAM). 
Depending on the model, a training run lasts between 1 and 5 hours. We provide model hyper-parameters in Table~\ref{tab:model_hyperparameters} and the hyper-parameters of the encoders in Table~\ref{tab:encoders_hyperparameters}.

In order to stabilize our training procedure, expression values were processed in two steps: first, we use a log-transform ($val = \log(1+val)$) to minimize the effect of outliers; secondly, we normalized expression values per tissue as it has been shown to stabilize training when using mean squared error loss.

%For all trainings using the cross-attention aggregation, the number of heads is set to 1. For the Perceiver Resampler, the number of heads is also set to 1, with 8 resampled tokens. The C-abstractor uses a kernel size of 3, resamples 8 tokens and has 1 residual layer. 

\begin{table}[h!]
  \caption{\small Model hyper-parameters}
  \label{tab:model_hyperparameters}
  \vspace{0.1cm}
  \centering
  { \small
  \begin{tabular}{l|c}
    \hline
    \textbf{Hyper-parameter} & \textbf{Value} 
    \\
    \hline
        Cross-Attn: number of heads  & 8 \\
        Perceiver Resampler: number of layers & 1 \\
        Perceiver Resampler: number of resampled tokens & 8 \\
        C-abstractor: kernel size & 3 \\
        C-abstractor: number of residual layers & 2 \\
        C-abstractor: number of resampled tokens & 8 \\
        Maximum number of nucleotides in \dna sequences (Enformer) & 196,608  \\
        Maximum number of nucleotides in \dna sequences (NT-v2) & 12,288 \\
        Maximum number of nucleotides in \rna sequences  & 12,288  \\
        Maximum number of amino-acids in protein sequences  & 1,200  \\
        \hline
  \end{tabular}
  }
\end{table}

\begin{table}[htbp!]
  \caption{\small Encoder hyper-parameters}
\label{tab:encoders_hyperparameters}
  \vspace{0.1cm}
  \centering
  { \small 
  \begin{tabular}{l|c}
    \hline
    \textbf{Hyper-parameter} & \textbf{Value} 
    \\
    \hline
        \nt: maximum number of tokens &  2,048 \\
        \nt: number of attention heads & 16\\
        \nt: embedding dimension & 768\\
        \nt: number of layers & 24\\
        \nt: activation & swish \\
        \enformer: number of parameters & 110M \\
        \enformer: embedding dimension & 1,536 \\
        \enformer: number of Transformer layers & 8 \\
        \textit{ESM-2-150M}: number of attention heads & 20 \\
        \textit{ESM-2-150M}: embedding dimension & 640 \\
        \textit{ESM-2-150M}: number of layers & 30 \\
        \hline
  \end{tabular}
  }
\end{table}
%Give more details about training procedure, seeds, hardware, training times etc.

\section{Experimental details}
\subsection{Attention maps analysis}
\label{sec:att_map_analysis}

Section \ref{sec:results_transfer} introduces significant results on \modelnamenospace's ability to leverage pre-trained modality-specific encoders. When finetuning the pre-trained encoders together, there is specialization on specific elements of the sequence, as observed in the \rna attention maps in Figure~\ref{fig:combined}-right. These maps have been obtained through the following process:
\begin{enumerate}
    \item Focusing on the pre-trained \rna encoder, we take the attention weights for each layer and head after running the \modelname finetuning process. For each layer and head, we compute how much attention (percentage) is directed towards a specific region of interest of the \rna sequence. The regions we consider are \textit{3UTR}, \textit{5UTR}, and \textit{CDS}, and we use the following equation to compute the exact ratio of attention:
    \begin{equation}
        \label{eq:attention_maps}
        \rho(f) = \frac{1}{\mathbf{X_{\text{rna}}}}\sum_{x\in\mathbf{X_{\text{rna}}}}\frac{\sum_i\sum_jf(i)\mathbf{1}(\alpha(i,j)>\mu)}{\sum_i\sum_j\mathbf{1}(\alpha(i,j)>\mu)}
    \end{equation}
    where $\mathbf{X_{\text{rna}}}$ is the set of \rna sequences in the test set, $\alpha(i,j)$ is the attention coefficient between tokens $i$ and $j$, $f(i)$ is an auxiliary function that equals 1 if token $i$ belongs to the region of interest in the sequence (e.g. \textit{3UTR}), and $\mu$ is a threshold value (we choose 0.01). We denote these attention maps by $\rho_{\text{IF}}$.
    \item We repeat this process for the finetuning run in which only the \rna encoder is used (\rna only in Table \ref{tab:omics_integration}). We denote these attention maps by $\rho_{\text{NT}}$.
    \item For each layer and head, we compute the ratio $\Delta\rho = (\rho_{\text{IF}} - \rho_{\text{NT}}) / \rho_{\text{NT}}$. For simplicity, we cap these values to 1 (i.e., attention rate is doubled in the \modelname case compared to finetuning a \rna encoder alone).
    \item In addition, for both $\rho_{\text{IF}}$ and $\rho_{\text{NT}}$, we can consider the samples in the test set $\mathbf{X_{\text{rna}}}$ and build a distribution of attention rates per element of interest. Comparing both distributions (the one coming from $\rho_{\text{IF}}$ and the one coming from $\rho_{\text{NT}}$), we can carry out a t-test per layer and head. We follow these t-tests and select the combinations of layers and heads in which there are statistically significant differences (i.e., p < 0.05) between $\rho_{\text{IF}}$ and $\rho_{\text{NT}}$.
    \item Figure~\ref{fig:combined}-right shows the ratio $\Delta\rho$ for those pairs of layers and heads in which statistically significant differences are observed. The rest is set to zero.
\end{enumerate}

\section{Additional results}
\label{appendix:additional_results}

\subsection{Full results over every tissue}

\begin{figure}[h]
\centering
\includegraphics[width = 0.6\textwidth]{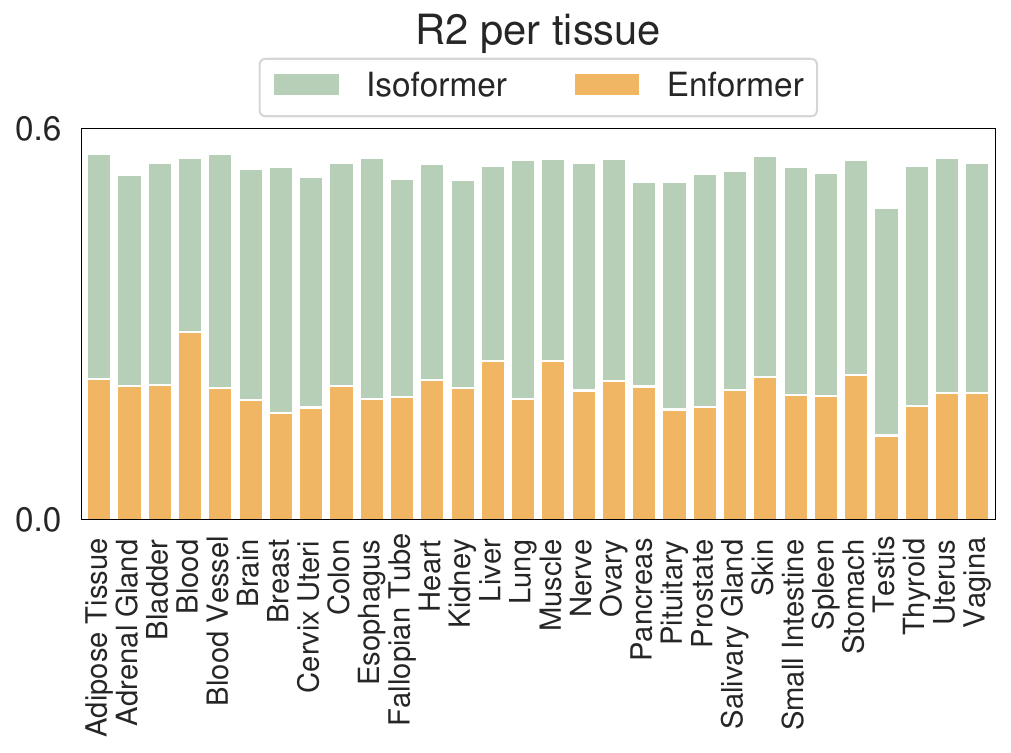}
\caption{\small Performance of \modelname and Enformer~\cite{avsec2021effective} per tissue on all tissues.}
\label{fig:r2_per_tissue_all}
\end{figure}

Figure~\ref{fig:r2_per_tissue_all} shows the performance of \modelname using \enformer as \dna encoder for each of the 30 tissues. \modelname outperforms the \dnanospace-only \enformer model on every tissue.

\clearpage

\end{document}